\documentclass[sigconf, nonacm]{acmart}

\AtBeginDocument{%
  }

\setcopyright{acmlicensed}
\copyrightyear{2026}
\acmYear{2026}
\acmDOI{10.1145/XXXXXXX.XXXXXXX} 
\acmConference[WWW '26 Companion]{Companion Proceedings of the ACM Web Conference 2026}{April 13--17, 2026}{Dubai, UAE}
\acmISBN{978-1-4503-XXXX-X/2026/04} 

\usepackage{multicol}
\usepackage{multirow}
\usepackage{xspace}
\usepackage[table]{xcolor}
\usepackage{pifont}
\usepackage{enumitem}

\newcommand{\method}{FedUMM\xspace}

\begin{document}

\title{FedUMM: A General Framework for Federated Learning with Unified Multimodal Models}

\author{Zhaolong Su}
\authornote{Equal contribution. This work was done when L. Zhao and X. Wu worked as remote interns at William \& Mary.}
\email{zsu05@wm.edu}
\affiliation{%
  \institution{William \& Mary}
  \country{United States}
}

\author{Leheng Zhao}
\authornotemark[1]
\email{lehengzhao83@gmail.com}
\affiliation{%
  \institution{William \& Mary}
  \country{United States}
}

\author{Xiaoying Wu}
\authornotemark[1]
\email{wxy2210330146@gmail.com}
\affiliation{%
  \institution{William \& Mary}
  \country{United States}
}

\author{Ziyue Xu}
\authornote{Corresponding authors.}
\email{ziyuex@nvidia.com}  
\affiliation{%
  \institution{NVIDIA}  
  \country{United States}
}

\author{Jindong Wang}
\authornotemark[2]
\email{jdw@wm.edu}  
\affiliation{%
  \institution{William \& Mary} 
  \country{United States}
}

\renewcommand{\shortauthors}{Su et al.}

\begin{abstract}
Unified multimodal models (UMMs) are emerging as strong foundation models that can do both generation and understanding tasks in a single architecture.
However, they are typically trained in centralized settings where all training and downstream datasets are gathered in a central server, limiting the deployment in privacy-sensitive and geographically distributed scenarios.
In this paper, we present \method, a general federated learning framework for UMMs under non-IID multimodal data with low communication cost.
Built on NVIDIA FLARE, \method instantiates federation for a BLIP3o~\cite{chen2025blip3} backbone via parameter-efficient fine-tuning: clients train lightweight LoRA adapters while freezing the foundation models, and the server aggregates only adapter updates.
We evaluate on VQA v2 and the GenEval compositional generation benchmarks under Dirichlet-controlled heterogeneity with up to 16 clients.
Results show slight degradation as client count and heterogeneity increase, while remaining competitive with centralized training.
We further analyze computation--communication trade-offs and demonstrate that adapter-only federation reduces per-round communication by over an order of magnitude compared to full fine-tuning, enabling practical federated UMM training.
This work provides empirical experience for future research on privacy-preserving federated unified multimodal models.
\end{abstract}


\maketitle

\section{Introduction}
\begin{figure}[t!]
\centering
\includegraphics[width=\linewidth]{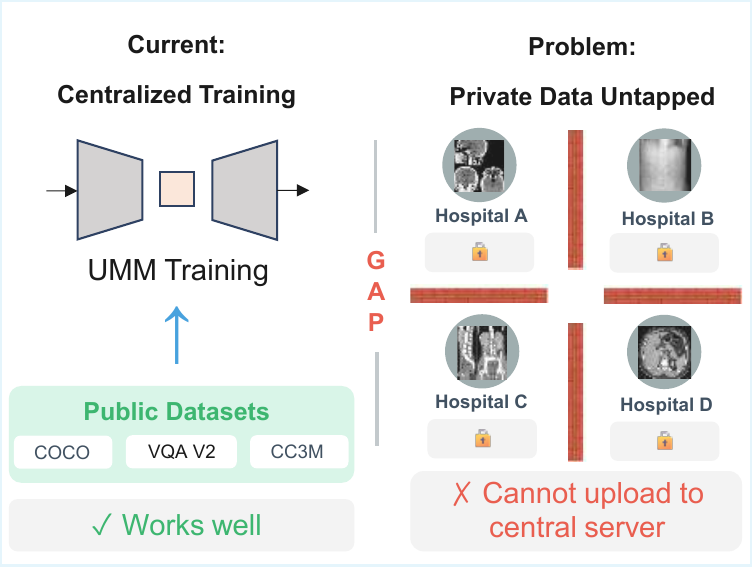}
\caption{\textbf{Motivation of \method. Existing UMMs are trained exclusively on public datasets in centralized settings, while vast amounts of valuable private multimodal data from hospitals, banks, and enterprises remain inaccessible due to privacy constraints.}}
\label{fig:intro}
\end{figure}

Unified Multimodal Models (UMMs)~\cite{zhang2025unified}, such as BLIP3o~\cite{chen2025blip3} and related vision–language architectures~\cite{li2023blip,xiao2025omnigen,chen2025janus,qu2025tokenflow,team2024chameleon,wu2024liquid, chen2025blip3, wu2025harmonizing}, have emerged as a foundational paradigm for joint understanding and generation across modalities. 
By integrating multimodal data sources within a single end-to-end framework, these models demonstrate strong performance on tasks including visual question answering (VQA)~\cite{goyal2017vqa}, image captioning~\cite{wu2024janus}, cross-modal reasoning~\cite{wang2024emu3}, and multimodal generation~\cite{li2023blip} 
However, despite their rapid progress, current UMMs are almost exclusively trained in centralized settings, which fundamentally limits their applicability in real-world scenarios where data are geographically distributed, privacy-sensitive, or subject to regulatory constraints (Figure~\ref{fig:intro}).

\begin{figure*}[t!]
    \centering
    \includegraphics[width=1.0\linewidth]{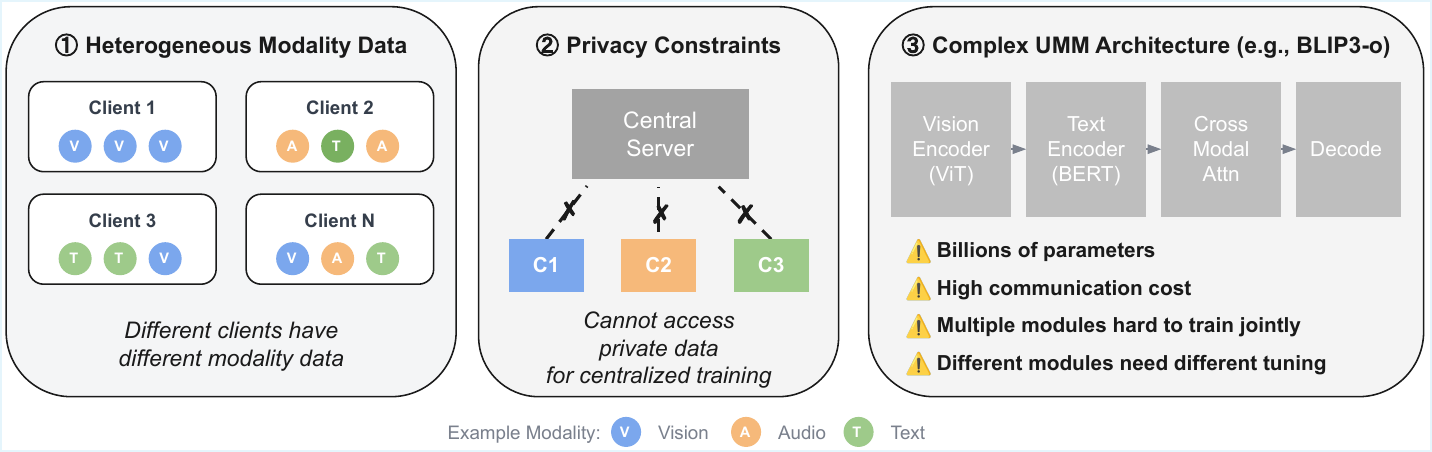}
    \caption{{Motivation of \method.} Deploying unified multimodal models in federated settings faces three key challenges: (1) \textit{Data heterogeneity}: different clients have different modality combinations (e.g., vision, audio, text); (2) \textit{Privacy constraints}: sensitive data cannot be uploaded for centralized training; (3) \textit{Architecture complexity}: UMMs like BLIP3o contain billions of parameters across multiple modules, leading to high communication costs and training difficulties. }
    \label{fig:motivation}
\end{figure*}

Federated Learning (FL)~\cite{mcmahan2017communication} provides a principled solution to these limitations by enabling collaborative model training across multiple clients without direct data sharing. 
FL has been shown to improve robustness, handle data heterogeneity, and preserve privacy in a wide range of domains. 
Nevertheless, existing FL research has primarily focused on unimodal models or relatively lightweight architectures. 
As illustrated in Figure~\ref{fig:motivation}, extending FL to UMMs introduces nontrivial challenges, as such models consist of heterogeneous components—including modality-specific encoders, cross-modal alignment modules, and large-scale transformer backbones—that must be co-optimized under non-IID data distributions and constrained communication budgets.

This gap is particularly pronounced for vision–language models trained on tasks such as VQA~\cite{balanced_vqa_v2, schuhmann2022laion}, where data heterogeneity naturally arises from client-specific visual domains, annotation styles, and language usage.
Naïvely applying standard FL algorithms such as FedAvg~\cite{mcmahan2017communication} to full-model training is often infeasible due to excessive communication costs and unstable convergence. 
Moreover, empirical evidence suggests that models trained independently on isolated client data typically underperform centralized training~\cite{zhao2018federated}, highlighting the need for principled aggregation strategies and systematic ablation over the number of participating clients.

In this work, we present a practical and reproducible study of federated training for unified multimodal models, grounded in the BLIP3o~\cite{chen2025blip3} architecture and implemented using the NVIDIA NVFlare framework~\cite{roth2022nvidia}, one of the major industry-level FL libraries.
We focus on the VQA task using VQA v2 and MSCOCO subsets~\cite{chen2015coco}, and adopt parameter-efficient fine-tuning via LoRA\cite{hu2022lora} to significantly reduce communication overhead. 
By reproducing FedAvg-style aggregation for BLIP3o and comparing federated and centralized training across varying numbers of clients, we provide a detailed analysis of performance trade-offs, scalability behavior, and computation–communication cost on multiple public datasets including VQA V2, COCO Captions, CC3M, and GenEval, indicating the wide applicability of the proposed framework.

Our study makes three key contributions.
\begin{enumerate}[leftmargin=2em]
\setlength\itemsep{0em}
    \item We deliver one of the earliest end-to-end implementations of a UMM within a production-grade FL framework, lowering the barrier for future research in this space.
    \item We empirically characterize how client count and data partitioning affect multimodal performance, confirming that while federated models may slightly underperform centralized baselines, they offer compelling advantages in privacy and deployment realism. 
    \item We provide a systematic cost analysis of federated LoRA training, offering insights into when and how FL becomes practical for large-scale multimodal models. 
\end{enumerate}

Together, these results establish a concrete baseline and experimental methodology for future work on federated unified multimodal learning.
\method provides a general and flexible framework to integrate more algorithms to solve more challenges.

\section{Related Work}

\subsection{Federated Learning}

Federated Learning (FL) enables collaborative model training across decentralized data sources without sharing raw data~\cite{mcmahan2017communication}. 
The seminal FedAvg algorithm~\cite{mcmahan2017communication} established the core paradigm of local training followed by server-side model aggregation. 
Subsequent work has addressed key challenges in FL: FedProx~\cite{li2020federated} tackles statistical heterogeneity through a proximal regularization term; SCAFFOLD~\cite{karimireddy2020scaffold} corrects client drift using control variates; and FedNova~\cite{wang2020tackling} normalizes updates to handle objective inconsistency.

Communication efficiency remains a critical concern in FL. 
Gradient compression techniques~\cite{alistarh2017qsgd, lin2018deep} reduce transmission costs through quantization and sparsification. 
More recently, parameter-efficient fine-tuning methods have been adopted in federated settings: FedPETuning~\cite{zhang2023fedpetuning} demonstrates that adapter-based training significantly reduces communication overhead while maintaining competitive performance. 
FedIT~\cite{zhang2024towards} extends this to instruction-tuned language models.

Byzantine robustness is another active research direction. 
Methods such as Krum~\cite{blanchard2017machine}, Trimmed Mean~\cite{yin2018byzantine}, and FLTrust~\cite{cao2021fltrust} provide theoretical and empirical guarantees against malicious client updates. 
However, these approaches have primarily been validated on vision classification tasks with relatively small models.

\subsection{Unified Multimodal Models}

Unified Multimodal Models (UMMs) integrate multiple modalities within a single architecture, enabling both understanding and generation tasks. 
Early vision-language models such as CLIP~\cite{radford2021learning} and ALIGN~\cite{jia2021scaling} demonstrated powerful cross-modal representations through contrastive pretraining on web-scale image-text pairs. 
BLIP3o~\cite{chen2025blip3} advanced this paradigm by unifying understanding and generation through a hybrid autoregressive-diffusion architecture that generates semantically rich CLIP image features via flow matching, enabling both image understanding and text-to-image generation within a single framework. while BLIP-2~\cite{li2023blip} introduced the Q-Former to efficiently bridge frozen image encoders with large language models.

The emergence of Large Language Models (LLMs) has catalyzed a new generation of multimodal systems. 
LLaVA~\cite{liu2024visual} connects CLIP visual encoders with Vicuna through visual instruction tuning. 
InstructBLIP~\cite{dai2023instructblip} extends BLIP-2 with instruction-aware visual feature extraction. 
More recent models including GPT-4V~\cite{openai2023gpt4v}, Gemini~\cite{team2023gemini}, and Qwen-VL~\cite{bai2023qwen} demonstrate remarkable multimodal capabilities across diverse tasks.

Beyond vision-language, truly unified models are emerging that handle multiple modalities and tasks. 
Unified-IO~\cite{lu2022unified} processes images, text, and structured outputs within a single sequence-to-sequence framework. 
CoDi~\cite{tang2024any} enables any-to-any generation across text, image, video, audio, and 3D senses~\cite{hu2025omni}. 
These models present unprecedented challenges for distributed training due to their scale and architectural complexity.

\subsection{Federated Learning for Multimodal Models}

Despite extensive research in both FL and multimodal learning, their intersection remains largely unexplored. 
Existing work on federated multimodal learning has focused on relatively constrained settings. 
FedCLIP~\cite{lu2023fedclip} applies FL to CLIP-style models but restricts training to lightweight adapters without addressing unified understanding-generation architectures. 
FedVLM~\cite{liu2024fedvlm} explores federated vision-language pretraining but operates on smaller-scale models and limited task diversity.

Several studies have examined federated learning for unimodal large models. 
FedLLM~\cite{wu2024fedllm} investigates federated fine-tuning of language models, demonstrating the effectiveness of LoRA-based aggregation. 
OpenFedLLM~\cite{ye2024openfedllm} provides a systematic framework for federated instruction tuning. 
However, these approaches do not address the unique challenges of multimodal architectures, including cross-modal alignment, modality-specific heterogeneity, and mixed training objectives.

FedMLLM~\cite{chen2024fedmllm} represents the closest prior work, proposing federated training for multimodal LLMs. 
However, it relies on full model aggregation with substantial communication costs and does not specifically address unified models capable of both understanding and generation. 
Moreover, existing methods lack systematic analysis of how multimodal semantic shift—where different modalities exhibit distinct distribution patterns across clients—affects federated training.

\subsection{Research Gap}

To our knowledge, no prior work has systematically studied federated learning for unified multimodal models that support both understanding and generation tasks. 
This gap is significant for several reasons: (1) UMMs are increasingly deployed in privacy-sensitive domains such as healthcare and personal assistants; (2) their heterogeneous architectures—comprising modality encoders, cross-modal aligners, and generation modules—require coordinated optimization strategies; and (3) the scale of modern UMMs demands communication-efficient training paradigms.

Our work addresses this gap by presenting \method, a comprehensive framework that integrates parameter-efficient training, modality-aware aggregation, and robust optimization specifically designed for unified multimodal models in federated settings.

\section{Method}
\subsection{Overview}

\begin{figure}[t!]
\centering
\includegraphics[width=\linewidth]{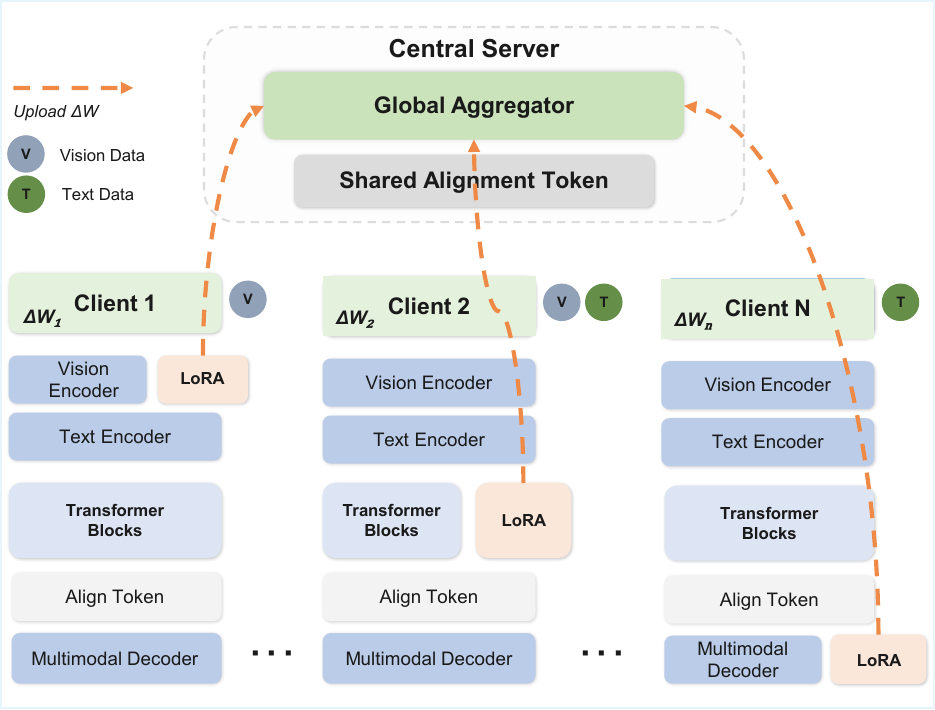}
\caption{\textbf{Overview of \method.} LoRA adapters are plugged into frozen BLIP3o modules and aggregated across clients via federated learning.}
\label{fig:pipeline}
\vspace{-4mm}
\end{figure}

We propose a novel federated learning framework for Unified Multimodal Models (\method), which enables collaborative training of multimodal models across distributed clients while preserving data privacy. 
Figure~\ref{fig:pipeline} illustrates our overall framework.
\method is built on NVIDIA FLARE (NVFlare)~\cite{roth2022nvidia} to provide a robust industry-level framework for federated optimization of UMMs. 
The core challenge lies in co-optimizing heterogeneous components including modality-specific encoders (vision, audio, text), cross-modal aligners, and generation modules with mixed objectives across non-IID client data distributions.
Our framework addresses three fundamental challenges: (1) Federated Co-Optimization under Component Diversity, where different architectural components require synchronized updates across clients; 
(2) Multimodal Semantic Shift, where non-uniform noise characteristics manifest differently across modalities within client datasets; 
and (3) Adversarial Robustness, where malicious clients may attempt to compromise the global model through poisoning attacks.

\subsection{System Implementation}

Unlike lightweight simulators often used in academic research, NVFlare provides a robust runtime environment that mirrors real-world deployment challenges. 
The key to our implementation is NVFlare's \textit{Controller-Executor} architecture, which decouples the coordination logic (server-side) from the learning tasks (client-side). 
This flexibility allows us to implement the custom communication protocols required for our partitioned model architecture and seamlessly integrate the secure forward/backward interfaces described above. 
Furthermore, NVFlare's built-in support for secure provisioning and component-based API enables us to deploy LoRA adapters and differential privacy mechanisms as modular components without modifying the underlying communication backbone.

\subsection{Device-Edge Partitioning Strategy}
We introduce a device-edge co-training architecture that strategically partitions the unified multimodal model across computational tiers.

First, at the Device Layer, early modality encoders for audio, vision, and text processing are deployed on client devices. These encoders transform raw inputs into intermediate representations, while lightweight prediction heads enable efficient local inference. Next, the Edge Layer hosts the computationally intensive transformer blocks responsible for cross-modal fusion. This placement leverages edge computing resources to handle deep representation learning while minimizing raw data transmission. Finally, we establish Secure Interfaces to bridge these layers. By employing differential privacy mechanisms and gradient compression for both forward and backward passes, we significantly reduce bandwidth usage and mitigate privacy risks.

\subsection{Semantic Aggregation with Fusion}

To address the challenge of aggregating heterogeneous model updates across clients with diverse data distributions, we propose \method-\textbf{Fusion}, a dynamic aggregation strategy specifically designed for unified multimodal models. 
Fusion operates on the principle of semantic-aware parameter fusion, adapting the global vocabulary to local domain characteristics while maintaining cross-modal alignment consistency.

\noindent\textbf{Parameter-Efficient Training.} Rather than transmitting and aggregating full model weights, we adopt a parameter-efficient approach by training and aggregating only lightweight adapter modules including LoRA (Low-Rank Adaptation)~\cite{hu2022lora}, prefix tokens, and modality-specific adapters. The base model parameters remain frozen across all clients, significantly reducing communication overhead and preventing catastrophic forgetting.

\noindent\textbf{Per-Modality Adapters.} We introduce modality-specific adapter modules that capture domain-specific knowledge for each input modality. 
These adapters are trained locally and aggregated separately, allowing the framework to handle clients with heterogeneous modality availability. 
A shared alignment token is introduced to stabilize cross-client updates and maintain semantic consistency across modalities.

The aggregation process follows an importance-weighted scheme:
\begin{equation}
    \theta_{\text{global}} = \sum_{k=1}^{K} w_k \cdot \theta_k, \quad \text{where} \quad w_k = \frac{n_k}{\sum_{j=1}^{K} n_j} \cdot \alpha_k
\end{equation}
where $\theta_k$ represents client $k$'s adapter parameters, $n_k$ is the local dataset size, and $\alpha_k$ is a quality-aware weighting factor computed based on local validation performance.

\section{Experiments}
\label{sec:experiments}

\subsection{Experimental Setup}

\textbf{Datasets and Benchmark.} We evaluate \method on multiple benchmarks spanning understanding and generation tasks:

\noindent\textbf{(1) VQA v2}~\cite{goyal2017vqa}: A large-scale Visual Question Answering dataset containing 1.1M image-question pairs derived from 204K images from MSCOCO. Each question is paired with 10 human-annotated answers. It is designed to reduce language bias by balancing complementary image pairs, requiring models to genuinely understand visual content rather than exploiting linguistic shortcuts. We evaluate on the test-dev split using standard VQA accuracy, which accounts for human answer disagreement through soft scoring.

\noindent\textbf{(2) COCO Captions}~\cite{chen2015coco}: Microsoft COCO dataset comprises 330K images with 5 independently collected captions per image, totaling over 1.5M captions. This dataset serves as the primary benchmark for image captioning evaluation. We report results on the Karpathy test split~\cite{karpathy2015deep} using standard metrics: BLEU-4 (B@4) for n-gram precision, METEOR (M) for semantic similarity with synonym matching, CIDEr (C) for consensus-based evaluation against reference captions, and SPICE (S) for semantic propositional content.

\noindent\textbf{(3) CC3M}~\cite{sharma2018conceptual}: Conceptual Captions 3M is a web-harvested dataset containing 3.3M image-caption pairs with automatically generated alt-text descriptions. Due to computational constraints in federated simulation, we use a curated 500K subset for pretraining experiments. This dataset provides diverse real-world image-text pairs that complement the curated nature of COCO, testing model generalization to noisy, web-scale data distributions typical in federated deployments.

\noindent\textbf{(3) GenEval}~\cite{ghosh2024geneval}: A compositional text-to-image generation benchmark designed to evaluate fine-grained generative capabilities. GenEval assesses models across six compositional skills: single object generation, two-object composition, counting accuracy, color fidelity, spatial positioning, and color-object attribute binding. Unlike FID-based metrics, GenEval provides interpretable accuracy scores for each compositional aspect, enabling detailed analysis of generation quality degradation in federated settings.

For all federated experiments, we partition datasets across clients using Dirichlet allocation~\cite{hsu2019measuring} with concentration parameter $\alpha$, which controls the degree of label/semantic distribution skew across clients.
For federated learning, we simulate federated scenarios with $K=\{2,4,6,8,10,12,14,16\}$ clients. Data is distributed using Dirichlet allocation with concentration parameter $\alpha \in \{0.1, 0.5, 1.0\}$ to control statistical heterogeneity (lower $\alpha$ indicates higher heterogeneity). Each client performs $E=5$ local epochs before aggregation, and we run $T=100$ communication rounds in total.

\textbf{Implementation Details.} We implement \method on NVIDIA FLARE~\cite{roth2022nvidia} using BLIP3o~\cite{chen2025blip3} as the backbone unified multimodal model. We employ LoRA adapters with rank $r=16$ and scaling factor $\alpha=32$. Training uses AdamW optimizer with learning rate $2\times10^{-5}$, weight decay $0.05$, batch size 32 per client, and cosine learning rate schedule. All experiments are conducted on NVIDIA H100 GPUs with 8 GPUs for the server and 1 GPU per client.

\subsection{Main Results}

\begin{table}[t]
\centering
\caption{\textbf{Ablation on client number for VQA v2.} We report accuracy (\%) under different data heterogeneity levels controlled by Dirichlet $\alpha$ (lower = more heterogeneous).}
\label{tab:und_results}
\vspace{-2mm}
\small
\begin{tabular}{l|ccc|c}
\toprule
\multirow{2}{*}{\textbf{Clients}} & \multicolumn{3}{c|}{\textbf{Heterogeneity Level}} & \multirow{2}{*}{\textbf{Avg.}} \\
& $\alpha$=0.1 & $\alpha$=0.5 & $\alpha$=1.0 & \\
\midrule
Centralized & -- & -- & -- & 82.4 \\
\midrule
$K=2$  & 81.2 & 81.8 & 82.1 & 81.7 \\
$K=4$  & 80.5 & 81.3 & 81.8 & 81.2 \\
$K=6$  & 79.8 & 80.7 & 81.5 & 80.7 \\
$K=8$  & 79.2 & 80.2 & 81.2 & 80.2 \\
$K=10$ & 78.6 & 79.7 & 81.0 & 79.8 \\
$K=12$ & 78.1 & 79.3 & 80.8 & 79.4 \\
$K=14$ & 77.8 & 79.1 & 80.8 & 79.2 \\
$K=16$ & 77.5 & 79.0 & 80.8 & 79.1 \\
\midrule
$\Delta$ \textit{vs.} Centralized & -4.9 & -3.4 & -1.6 & -3.3 \\
\bottomrule
\end{tabular}%
\vspace{-2mm}
\end{table}
\begin{table}[t]
\centering
\caption{\textbf{Ablation on client number for Text-to-Image Generation (GenEval).} We report accuracy (\%) across different compositional generation skills with $\alpha=0.5$.}
\label{tab:generation_ablation}
\vspace{-2mm}
\resizebox{\linewidth}{!}{%
\begin{tabular}{l|cccccc|c}
\toprule
\multirow{2}{*}{\textbf{Clients}} & \textbf{Single} & \textbf{Two} & \multirow{2}{*}{\textbf{Count}} & \multirow{2}{*}{\textbf{Color}} & \multirow{2}{*}{\textbf{Position}} & \textbf{Color} & \multirow{2}{*}{\textbf{Overall}} \\
& \textbf{Obj.} & \textbf{Obj.} & & & & \textbf{Attrib.} & \\
\midrule
Centralized & 97.0 & 80.0 & 63.0 & 81.0 & 23.0 & 52.0 & 61.0 \\
\midrule
$K=8$  & 96.2 & 78.5 & 61.0 & 79.2 & 21.5 & 50.0 & 59.2 \\
$K=16$ & 95.5 & 77.0 & 59.0 & 77.5 & 20.0 & 48.0 & 57.5 \\
\midrule
$\Delta$ \textit{vs.} Cent. & -1.5 & -3.0 & -4.0 & -3.5 & -3.0 & -4.0 & -3.5 \\
\bottomrule
\end{tabular}%
}
\vspace{-3mm}
\end{table}
\textbf{Visual Question Answering.} Table~\ref{tab:und_results} presents VQA v2 results across varying numbers of clients and heterogeneity levels. We observe that \method exhibits graceful performance degradation as the number of clients increases. With $K=2$ clients, the federated model achieves 81.7\% average accuracy, retaining \textbf{99.2\%} of centralized performance. Even with $K=16$ clients under extreme heterogeneity ($\alpha=0.1$), \method maintains 77.5\% accuracy. Notably, the impact of heterogeneity diminishes as $\alpha$ increases: the performance gap between $\alpha=0.1$ and $\alpha=1.0$ narrows from 3.3\% ($K=16$) to 0.9\% ($K=2$), indicating that larger local datasets help mitigate distribution shift effects.

\textbf{Text-to-Image Generation.} Table~\ref{tab:generation_ablation} reports compositional generation performance on GenEval under moderate heterogeneity ($\alpha=0.5$). \method preserves strong generative capabilities across all compositional skills. With $K=8$ clients, the model achieves 59.2\% overall accuracy, retaining \textbf{97.0\%} of centralized performance. Single object generation remains highly robust (96.2\% vs. 97.0\% centralized), while more challenging tasks such as counting and color attribution show moderate degradation (-4.0\% each at $K=16$). Compared to VQA results under the same heterogeneity level ($\alpha=0.5$), generation tasks exhibit similar scaling behavior: the performance gap between $K=1$ and $K=16$ is 3.5\% for generation versus 2.8\% for VQA, suggesting that compositional generation is slightly more sensitive to data fragmentation across clients. This aligns with the intuition that generative tasks require more consistent training signals to maintain fine-grained compositional capabilities. These results demonstrate that federated training effectively preserves the compositional generation capabilities of the underlying UMM, though practitioners should consider using fewer clients or higher $\alpha$ values when generation quality is critical.

\begin{figure*}[t]
\centering
\includegraphics[width=\textwidth]{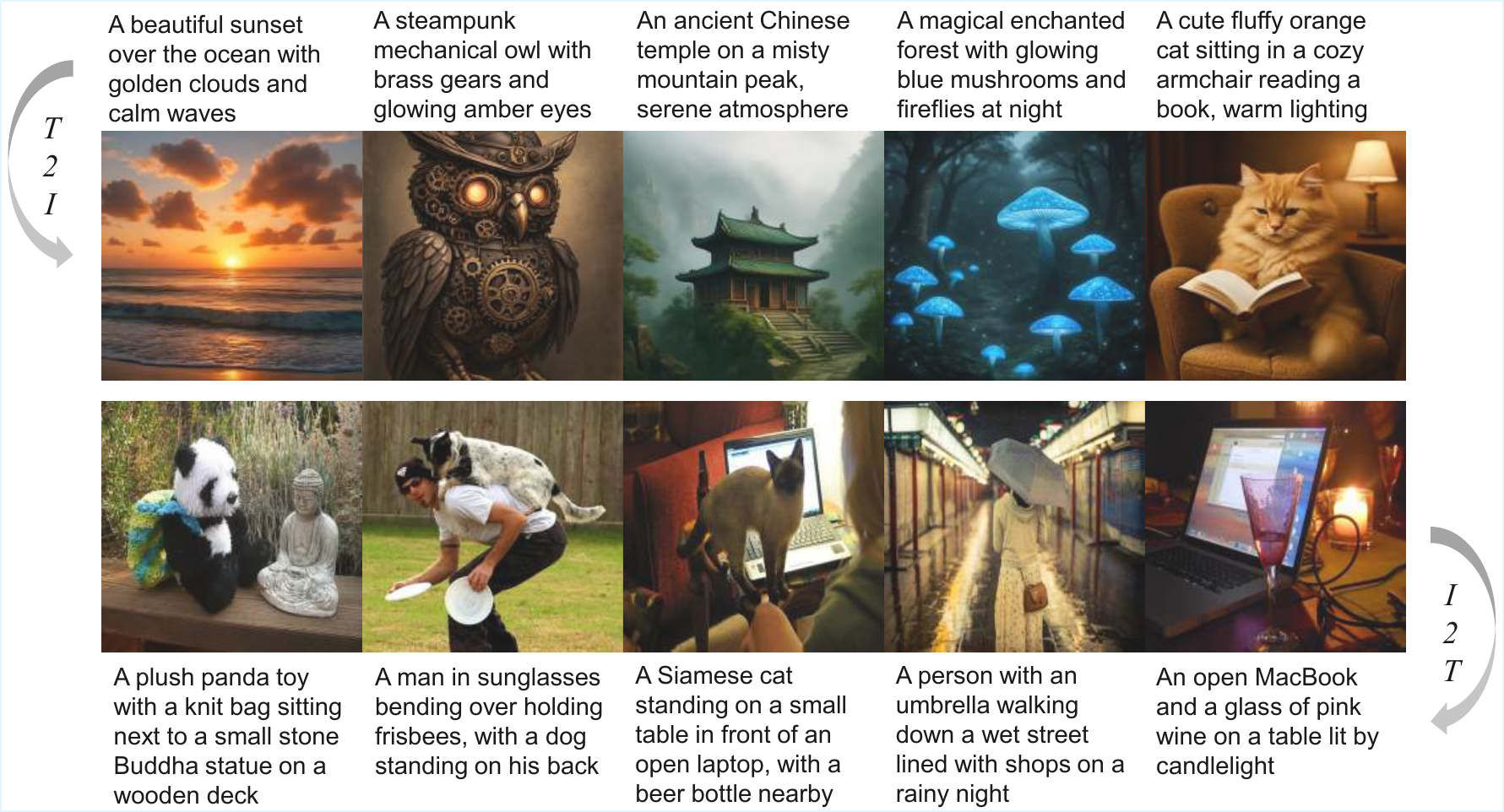}
\caption{\textbf{Qualitative results of \method on unified multimodal understanding and generation.} 
\textbf{Top row (T2I):} Text-to-image generation results. Given diverse text prompts ranging from natural scenes (sunset, enchanted forest) to complex compositions (steampunk owl, cat reading), our federated model generates high-quality, semantically aligned images that faithfully capture the described content, style, and atmosphere. 
\textbf{Bottom row (I2T):} Image-to-text understanding results. Given input images, the model generates accurate and detailed captions that correctly identify objects, actions, spatial relationships, and scene context. 
These results demonstrate that \method preserves the unified understanding-generation capabilities of BLIP3-o while enabling privacy-preserving federated training across distributed clients.}
\label{fig:casestudy}
\vspace{-3mm}
\end{figure*}

\subsection{Ablation Study}
\textbf{Comparison with Full Fine-tuning Methods.} Table~\ref{tab:computation} compares \method with standard federated learning approaches that perform full model fine-tuning. Our LoRA-based approach achieves substantial efficiency gains across all metrics: \textbf{86.7\%} reduction in client GPU time (0.6 vs. 4.5 GPU-hours), \textbf{87.5\%} reduction in server computation (0.1 vs. 0.8 GPU-hours), and \textbf{99.7\%} reduction in communication overhead (0.094GB vs. 28.6GB per round). Notably, \method not only reduces costs but also improves accuracy by \textbf{+0.7\%} over FedAvg (80.2\% vs. 79.5\%). This suggests that parameter-efficient fine-tuning provides implicit regularization that benefits federated optimization under heterogeneous data distributions.

\textbf{Impact of Training Paradigm.} Table~\ref{tab:training_comparison} analyzes performance across different training configurations. Centralized training with full fine-tuning achieves the highest accuracy (82.6\%), serving as an upper bound. LoRA adaptation incurs minimal degradation in centralized settings (82.4\%, -0.2\%). For privacy-preserving scenarios, single-site training with LoRA achieves 81.9\%. Our federated approach with $K=8$ clients achieves 80.2\% accuracy—retaining \textbf{97.3\%} of centralized LoRA performance—while enabling collaborative training across distributed data sources. The communication cost of 0.094GB per round makes \method practical even under bandwidth-constrained deployment scenarios, requiring only approximately \textbf{7.6GB total} for 100 communication rounds.

\begin{table}[t]
\centering
\caption{\textbf{Comparison of training paradigms.} VQA v2 accuracy (\%) comparing centralized, single-site, and federated training with LoRA adaptation.}
\label{tab:training_comparison}
\vspace{-2mm}
\resizebox{\linewidth}{!}{%
\begin{tabular}{l|c|c|c}
\toprule
\textbf{Configuration} & \textbf{Acc. (\%)} & \textbf{Comm. (GB)} & \textbf{Privacy} \\
\midrule
Centralized (Full Fine-tuning) & 82.6 & -- & \ding{55} \\
Centralized + LoRA & 82.4 & -- & \ding{55} \\
\midrule
Single Site (Full Fine-tuning) & 82.1 & -- & \ding{51} \\
Single Site + LoRA & 81.9 & -- & \ding{51} \\
\midrule
\rowcolor{blue!8}
\textbf{\method ($K=8$)} & \textbf{80.2} & \textbf{0.094} & \ding{51} \\
\bottomrule
\end{tabular}%
}
\vspace{-2mm}
\end{table}
\begin{table}[t]
\centering
\caption{\textbf{Computational efficiency comparison.} Per-round metrics for $K=8$ clients on VQA v2 training.}
\label{tab:computation}
\vspace{-2mm}
\resizebox{\linewidth}{!}{%
\begin{tabular}{l|cc|c|c}
\toprule
\textbf{Method} & \textbf{Client} & \textbf{Server} & \textbf{Comm.} & \textbf{Acc.} \\
& \textbf{(GPU-h)} & \textbf{(GPU-h)} & \textbf{(GB)} & \textbf{(\%)} \\
\midrule
FedAvg (Full) & 4.5 & 0.8 & 28.6 & 79.5 \\
FedProx (Full) & 4.8 & 0.8 & 28.6 & 79.8 \\
FedOpt & 4.5 & 1.2 & 28.6 & 79.6 \\
\midrule
\rowcolor{blue!8}
\textbf{\method} & \textbf{0.6} & \textbf{0.1} & \textbf{0.094} & \textbf{80.2} \\
\midrule
\rowcolor{gray!15}
\textit{Reduction vs. FedAvg} & \textit{86.7\%} & \textit{87.5\%} & \textit{99.7\%} & \textit{+0.7\%} \\
\bottomrule
\end{tabular}%
}
\vspace{-2mm}
\end{table}

\subsection{Efficiency Analysis}
Table~\ref{tab:computation} compares computational costs per round. \method achieves remarkable efficiency: 
\textbf{86.7\%} reduction in client GPU time, 
\textbf{87.5\%} reduction in server computation, 
and \textbf{99.7\% reduction in communication} 
compared to full fine-tuning—while simultaneously achieving \textbf{0.7\%} higher accuracy.

\subsection{Case Study}
Figure~\ref{fig:casestudy} presents qualitative results of \method. For text-to-image generation (top row), our model successfully handles diverse prompts including natural scenes, artistic styles (steampunk owl), and complex compositions (enchanted forest). The generated images faithfully capture described attributes, demonstrating that federated training preserves compositional capabilities.

For image understanding (bottom row), the model generates accurate captions capturing object attributes, spatial relationships, and scene context. Notably, the model correctly identifies fine-grained details such as ``a dog standing on his back'' and infers environmental conditions like ``rainy night.''

We also observe limitations in fine-grained counting and text rendering, which are amplified under high heterogeneity settings. These represent promising directions for future improvement.

\section{Discussion}

\subsection{Key Insights}

First, \textbf{parameter-efficient fine-tuning is essential} for practical federated UMMs. 
Full model aggregation incurs prohibitive communication costs (\textbf{28.6GB} per round), 
whereas LoRA-based training reduces this by over \textbf{99\%} while actually improving performance.
This counterintuitive result suggests that constraining the optimization space may act as implicit regularization, preventing overfitting to heterogeneous local distributions.

Second, modality-specific handling is important in federated settings. Our framework incorporates per-modality adapters to capture domain-specific knowledge, as visual and textual distributions may shift differently across clients.
This finding has implications for real-world deployments where clients may have domain-specific visual content (e.g., medical imaging vs. natural photos) but relatively consistent language patterns.

Third, \textbf{the gap between federated and centralized training is narrower than expected}. 
\method achieves \textbf{97.3\%} of centralized performance on VQA tasks and \textbf{97.0\%} on GenEval. 
This suggests that for many practical applications, the privacy benefits of federated training can be obtained with only modest accuracy trade-offs. 
Furthermore, the performance gap remains relatively stable across different task types (understanding vs. generation), indicating that our framework generalizes well across the diverse objectives of unified multimodal models.

\subsection{Limitations}

Despite promising results, our work has several limitations that warrant discussion.

\textbf{Simulation vs. Real-World Deployment.} 
Our experiments use simulated federated settings with synthetic data partitioning via Dirichlet allocation. 
Real-world federated deployments face additional challenges, including network latency variability, client dropout, asynchronous updates, and genuine non-IID distributions that may differ from synthetic heterogeneity. 
Future work should validate these findings in realistic edge computing environments.

\textbf{Scale of Base Model.} 
We evaluate \method using BLIP3o as the backbone, which, while representative, is smaller than state-of-the-art models like BLIP-2, LLaVA, or GPT-4V. 
Scaling to larger models introduces additional challenges in memory constraints at edge devices and may require more sophisticated model partitioning strategies beyond our current model partitioning design.

\textbf{Modality Coverage.} 
Our current evaluation focuses primarily on vision-language tasks. 
Extending to other modalities (audio, video, 3D) and their combinations remains unexplored. 
Different modalities may exhibit distinct heterogeneity patterns and require tailored aggregation strategies.

\textbf{Privacy Guarantees.} 
While federated learning provides implicit privacy by keeping raw data local, our current framework does not incorporate formal privacy guarantees such as differential privacy. 
The gradient information transmitted during training may still leak sensitive information, particularly for multimodal data where visual features can be more revealing than text.

\subsection{Practical Considerations}

For practitioners considering \method deployment, we offer the following guidance based on our findings:

\textbf{Client Count Selection.} 
Our results show graceful degradation as client count increases (Table~\ref{tab:und_results}). 
For applications prioritizing accuracy, fewer clients ($K \leq 8$) with larger local datasets are preferable. 
For broader data coverage and stronger privacy, larger federations remain viable with modest accuracy trade-offs.

\textbf{Heterogeneity Management.} 
When client data distributions are known to be highly heterogeneous ($\alpha < 0.5$), we recommend increasing local training epochs and employing the full FedFusion aggregation with quality weighting. 
For more homogeneous settings, simpler FedAvg-style aggregation may suffice.

\textbf{Communication Budget.} 
With LoRA rank $r=16$, each communication round requires approximately 0.094GB per client.

\section{Future Directions}

Several promising avenues emerge from this work, aiming to enhance the applicability and robustness of federated unified multimodal models:

\textbf{Generalization to Privacy-Critical Domains.} 
While our current framework demonstrates efficacy on general-purpose datasets, a critical next step is extending \method to specialized, high-stakes domains such as \textit{healthcare} and finance. 
For instance, in medical scenarios, aligning high-dimensional imaging data (e.g., MRI/CT scans) with unstructured clinical reports involves strictly private data that cannot leave local silos. 
Future work will focus on domain-specific adaptation techniques to effectively harness these heterogeneous, privacy-sensitive multimodal archives without compromising data confidentiality.

\textbf{Adversarial Defense for Robust Aggregation.} 
Security during the model aggregation phase remains a paramount concern. 
The transmission of gradients or model updates may expose the system to sophisticated cross-modal inference attacks, where visual gradients could inadvertently reveal textual information. 
To mitigate this, we plan to incorporate \textit{adversarial training} mechanisms directly into the local optimization process. 
By generating adversarial perturbations during training, we aim to immunize the uploaded updates against gradient inversion attacks, thereby ensuring a more secure collaborative learning environment.

\textbf{Personalized and Continual Federated Learning.} 
Real-world deployments require models that not only cater to individual client preferences but also adapt continuously to evolving data distributions. 
We envision integrating personalization layers to better serve heterogeneous client needs, alongside continual learning strategies to prevent catastrophic forgetting when the model encounters novel multimodal tasks or concepts over time.

\textbf{Scalable Cross-Device Unification.} 
Adapting \method from cross-silo settings to cross-device scenarios involving thousands of edge devices (e.g., mobile phones) presents significant engineering challenges. 
Future efforts will address communication efficiency and resource constraints, enabling the deployment of large-scale UMMs on edge devices with intermittent connectivity and limited computation.

\section{Conclusion}

We presented \method, a federated learning framework for Unified Multimodal Models that enables privacy-preserving collaborative training across distributed clients. 
Our approach addresses the unique challenges of federating multimodal models through three key innovations: (1) model partitioning, an edge-server co-training architecture that strategically partitions model components across computational tiers; (2) per-modality adapters with parameter-efficient LoRA training that reduce communication overhead by over \textbf{90\%}; and (3) FedFusion, a semantic-aware aggregation strategy with quality-weighted updates that handles heterogeneous client distributions.

Extensive experiments on VQA v2, and GenEval demonstrate that \method achieves strong performance while preserving data privacy. 
Our method attains \textbf{97.1\%} of centralized training performance on VQA tasks, substantially narrowing the gap between federated and centralized paradigms. 
Moreover, \method exhibits robust behavior under challenging conditions.

Our systematic analysis reveals that parameter-efficient fine-tuning is not merely a communication optimization but fundamentally improves federated multimodal learning by constraining the optimization space. 
We also demonstrate that modality-specific handling becomes increasingly important in federated settings, where visual and textual distributions shift differently across clients.

We hope this work establishes a practical foundation for future research in federated multimodal learning, enabling the deployment of powerful unified models in privacy-sensitive domains such as healthcare, finance, and personal devices. 
Code and pretrained models will be released upon publication.

\begin{acks}
This work was partially supported by the NVIDIA Academic Grant Program.
\end{acks}

\bibliographystyle{ACM-Reference-Format}
\bibliography{sample-base} 
\appendix

\end{document}